\documentclass[11pt,a4paper]{article}
\usepackage[hyperref]{emnlp-ijcnlp-2019}
\usepackage{times}
\usepackage{latexsym}

\usepackage{graphicx}
\usepackage{booktabs}
\usepackage{color}

\usepackage{url}
\usepackage{CJK}

\usepackage{enumitem}
\setenumerate[1]{itemsep=0pt,partopsep=0pt,parsep=\parskip,topsep=5pt}
\setdescription{itemsep=0pt,partopsep=0pt,parsep=\parskip,topsep=5pt}
\title{BiPaR: A Bilingual Parallel Dataset for Multilingual and Cross-lingual Reading Comprehension on Novels}

\aclfinalcopy

\author{Yimin Jing ,\ Deyi Xiong\thanks{\ \ Corresponding author} \and Yan Zhen \\ School of Computer Science and Technology, Soochow University, Suzhou, China \\ {\tt \{yymmjing,corezhen\}@gmail.com} \\ {\tt dyxiong@suda.edu.cn} \\}

\date{}

\date{}

\begin{document}
\maketitle
\begin{CJK}{UTF8}{gbsn}
\begin{abstract}
This paper presents \textit{BiPaR}, a \textbf{bi}lingual \textbf{pa}rallel novel-style machine \textbf{r}eading comprehension (MRC) dataset, developed to support multilingual and cross-lingual reading comprehension. The biggest difference between BiPaR and existing reading comprehension datasets is that each triple (Passage, Question, Answer) in BiPaR is written parallelly in two languages. We collect 3,667 bilingual parallel paragraphs from Chinese and English novels, from which we construct 14,668 parallel question-answer pairs via crowdsourced workers following a strict quality control  procedure. We analyze BiPaR in depth and find that BiPaR offers good diversification in prefixes of questions, answer types and relationships between questions and passages. We also observe that answering questions of novels requires reading comprehension skills of coreference resolution, multi-sentence reasoning, and understanding of implicit causality, etc.  With BiPaR, we build monolingual, multilingual, and cross-lingual MRC baseline models. Even for the relatively simple monolingual MRC on this dataset, experiments show that a strong BERT baseline is over 30 points behind human in terms of both EM and F1 score, indicating that BiPaR provides a challenging testbed for monolingual, multilingual and cross-lingual MRC on novels. The dataset is available at https://multinlp.github.io/BiPaR/.
\end{abstract}

\section{Introduction}

Machine reading comprehension is to evaluate how well computer systems understand natural language texts, where machines read a given text passage and answer questions about the passage. It has been regarded as a crucial technology for many applications such as question answering, dialogue systems \cite{nguyen2016ms, chen-etal-2017-reading, liu2018stochastic, wang2018multi} and so on. In order to enable machine to understand texts, large-scale reading comprehension datasets have been developed, such as CNN/Daily Mail \cite{hermann2015teaching}, SQuAD \cite{rajpurkar2016squad}, MS MACRO \cite{nguyen2016ms}, hotpotQA \cite{yang2018hotpotqa}, CoQA \cite{reddy2019coqa}, etc.

\begin{figure}[tt]
\centering
\includegraphics[height=4.3in, width=3.0in]{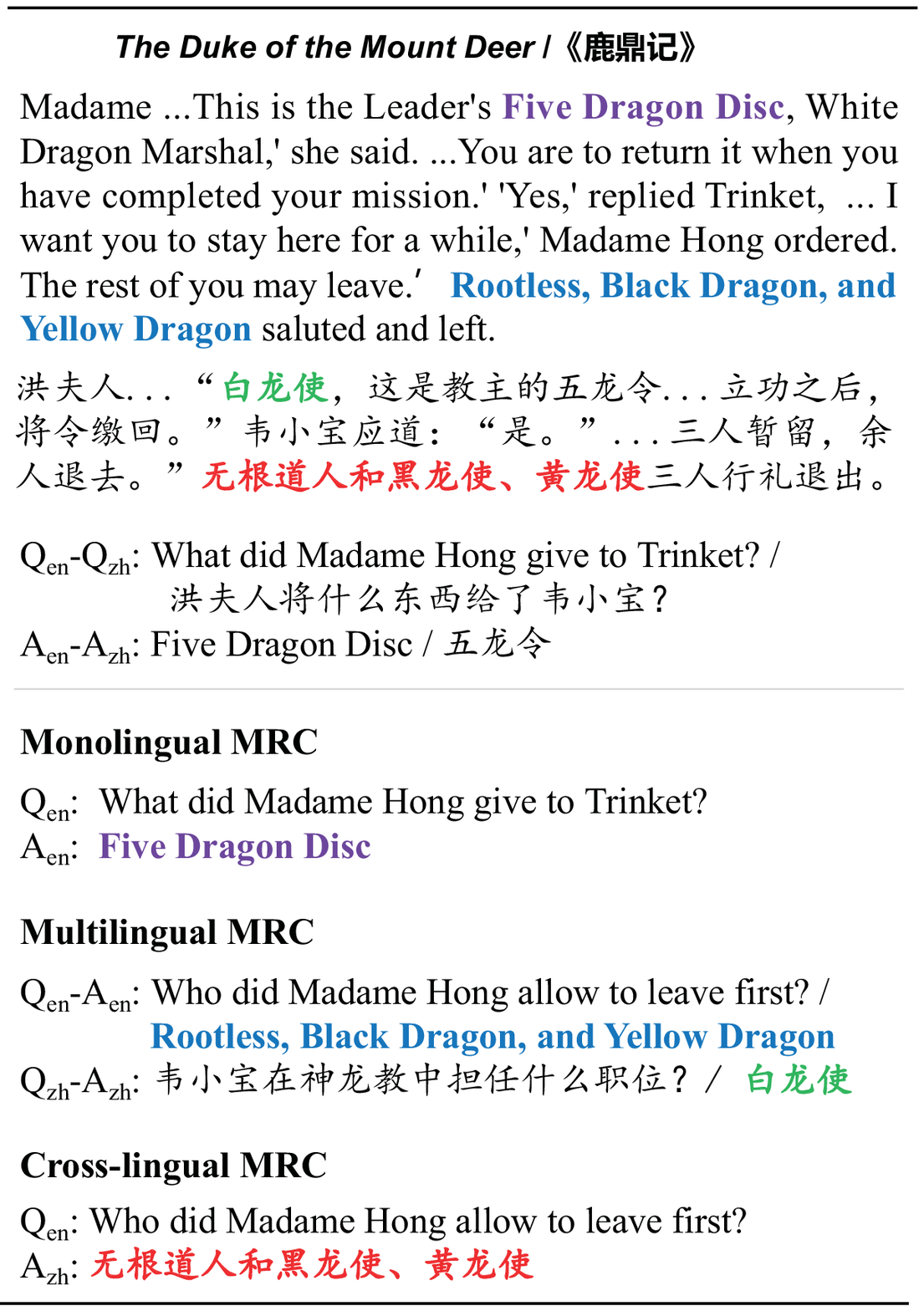}
\caption{Illustration of BiPaR with the monolingual, multilingual and cross-lingual MRC on the dataset.}
\label{sample}
\end{figure}

\begin{table*}[]
\small
\centering
\begin{tabular}{@{}lccll@{}}
\toprule
Dataset & \multicolumn{1}{l}{Language} & \multicolumn{1}{l}{Parallel} & Answer Type & Domain \\ \midrule
\begin{tabular}[c]{@{}l@{}}CNN/DM \cite{hermann2015teaching}\\ HLF-RC \cite{cui2016consensus}\\ SQuAD \cite{rajpurkar2016squad}\\ CMRC2018 \cite{cui2018span}\\ MS MACRO \cite{nguyen2016ms}\\ DuReader \cite{he2018dureader}\end{tabular} & \begin{tabular}[c]{@{}c@{}}EN\\ ZH\\ EN\\ ZH\\ EN\\ ZH\end{tabular} & \begin{tabular}[c]{@{}c@{}}${\times}$\\ ${\times}$\\ ${\times}$\\ ${\times}$\\ ${\times}$\\ ${\times}$\end{tabular} & \begin{tabular}[c]{@{}l@{}}Fill in entity\\ Fill in word\\ Span of words\\ Span of words\\ Manual summary\\ Manual summary\end{tabular} & \begin{tabular}[c]{@{}l@{}}News\\ Fairy/News\\ Wikipedia\\ Wikipedia\\ Web doc.\\ Web doc./CQA\end{tabular} \\
\midrule
BiPaR (this paper) & EN-ZH & ${\surd}$ & Span of words & \begin{tabular}[c]{@{}l@{}}Novels (Kongfu novels, \\ science fictions, etc.)\end{tabular} \\ \bottomrule
\end{tabular}
\caption{Comparison of BiPaR with several existing reading comprehension datasets.}
\label{Comparision}
\end{table*}

The majority of such datasets, unfortunately, are only for monolingual text understanding. To the best of our knowledge, there is no publicly available bilingual parallel reading comprehension dataset, which is exactly what BiPaR is mainly developed for, as illustrated in Figure \ref{sample}. BiPaR produces over 14K Chinese-English parallel questions from nearly 4K bilingual parallel novel passages with consecutive word spans from these passages as answers, following SQuAD \cite{rajpurkar2016squad}. Table \ref{Comparision} shows that BiPaR has two significant differences in comparison with existing datasets: (1) each (Passage, Question, Answer) triple is bilingual parallel, and (2) passages and questions are from novels. BiPaR's bilinguality and novel-based questions provide an interesting corner and unexplored territory for MRC. 

With an in-depth analysis on the manually created questions, we observe that answering these novel-style questions requires challenging skills: coreference resolution, multi-sentence reasoning, and understanding of implicit causality, etc. Further monolingual MRC experiments on BiPaR demonstrate that, the English BERT$\_$large model \cite{devlin2019bert} achieves an F1 score of 56.5\%, which is 35.4 points behind human performance (91.9\%), and the Chinese BERT$\_$Base model achieves an F1 score of 64.1\%, which is 28.0 points behind human performance (92.1\%), indicating there is a huge gap to be bridged for MRC on novels.

More interestingly, the bilinguality of BiPaR supports multilingual and cross-lingual MRC tasks on this dataset in addition to the traditional monolingual MRC. It is more cost-effective to build a single model that can handle machine reading comprehension on multiple languages, than building multiple MRC systems, one reading system for each language. Different from previous multilingual QA systems that are trained on independently developed datasets of different languages and domains, we are able to train a single multilingual MRC model to do MRC on different languages, by exploring BiPaR that is built parallelly on two different languages with alignments between triples of (Passage, Question, Answer) of the two languages, as shown in Figure \ref{sample}.

Yet another interesting task that we can do with BiPaR is cross-lingual reading comprehension. We define two types of cross-lingual MRC on BiPaR: (1) using questions in one language to find answers from passages written in another language and (2) finding answers from passages of two different languages for questions in one language.  The former is in essence similar to the early cross-lingual question answering (CLQA) \cite{aceves2008two, penas2009overview, perez2009information}. The intuitive approaches to CLQA are to translate the questions into the document's language \cite{sutcliffe2005cross, de2005miracle, aceves2007enhancing}, which, however, suffers from translation errors. The BiPaR dataset provides a potential opportunity for building cross-lingual MRC that does not rely on machine translation. 
 
To summarize, our contributions are threefold: 
\begin{itemize}
    \item We build the BiPaR, the first publicly available bilingual parallel dataset for MRC. The passages are novel paragraphs, originally written in Chinese or English and then translated into the other language. The questions are manually constructed undergoing a strict quality control procedure.
    \item We conduct an in-depth analysis on BiPaR, which reveals that MRC on novels is very challenging, requiring skills of coreference resolution, inter-sentential reasoning, implicit causality understanding, etc.
    \item We build monolingual, multilingual and cross-lingual MRC baseline models on BiPaR and provides baseline results as well as human performance on this dataset.
\end{itemize}

\section{Related Work}

\noindent\textbf{MRC Datasets and Models} Large-scale cloze-style datasets, such as CNN/Daily Mail \cite{hermann2015teaching}, have been automatically developed in the early days of MRC. Several neural network models have been proposed and tested on these datasets, such as ASReader \cite{kadlec2016text}, StanfordAttentiveReader \cite{chen2016thorough}, AoAReader \cite{cui2017attention}, etc. However, \citet{chen2016thorough} argue that such datasets may be noisy due to the automatic data creation method and co-reference errors. \citet{rajpurkar2016squad} propose SQuAD, a dataset created from English Wikipedia, where questions are manually generated by crowdsourced workers, and answers are spans in the Wikipedia passages. Along with the development of this dataset, a variety of neural MRC models have been proposed, such as BiDAF \cite{seo2016bidirectional}, R-NET \cite{wang2017gated}, ReasonNet \cite{shen2017reasonet}, DCN \cite{xiong2016dynamic}, QANet \cite{yu2018qanet}, SAN \cite{liu2018stochastic}, etc. Recent years have witnessed substantial progress made on this dataset. However, there are some limitations on SQuAD, which lie in that questions are created based on single passages, that answers are limited to a single span in passages, and that most questions can be answered from a single supporting sentence without requiring multi-sentence reasoning \cite{chen2018neural}. To address these limitations, a number of datasets have be built recently, such as MS MARCO \cite{nguyen2016ms}, DuReader \cite{he2018dureader}, TriviaQA \cite{joshi2017triviaqa}, RACE \cite{lai2017race}, NarrativeQA \cite{kocisky2018narrativeqa}, SQuAD2.0 \cite{rajpurkar2018know}, hotpotQA \cite{yang2018hotpotqa}, CoQA \cite{reddy2019coqa}, etc. These datasets and models are only for monolingual text understanding. By contrast, BiPaR, following these efforts of creating challenging MRC datasets, aims at setting up a new benchmark dataset for MRC on novels and bilingual/cross-lingual MRC. \\

\noindent{\bf Multilingual MRC and Datasets}  Previous studies on multilingual MRC are very limited. \citet{asai2018multilingual} propose a multilingual MRC system by translating the target language into a pivot language via runtime machine translation. They still rely on SQuAD to train the MRC model of the pivot language. No multilingual MRC dataset is created except that a Japanese and French test set is created by manually translating the test set of SQuAD into the two languages. \\

\noindent{\bf Multilingual/Cross-lingual QA}  Answering questions in multiple languages or retrieving answers from passages that are written in a language different from questions is an important capability for QA systems. To achieve this goal, QA@CLEF\footnote{http://www.clef-initiative.eu/track/qaclef} has organized a series of public evaluations for multilingual/cross-lingual QA \cite{magnini2006multilingual}. Widely-used approaches to multilingual/cross-lingual QA are to build monolingual QA systems and then adapt them to multilingual/cross-lingual settings via machine translation \cite{Lin2005CMUJS}. Such QA systems are prone to being affected by machine translation errors. Hence, various techniques have been proposed to reduce the errors of the machine translation module \cite{sutcliffe2005cross, de2005miracle, aceves2007enhancing}.

\section{Dataset Creation}
In this section, we elaborate on the three stages of our dataset creation process: collecting bilingual parallel passages, crowdsourcing question-answer pairs on those passages, and constructing multiple answers for the development and test set.

\subsection{Bilingual Parallel Passage Collection}

We select bilingual parallel passages from six Chinese and English novels with different topics, including Chinese martial arts, science fictions, fantasy literature, etc. These novels are either written in Chinese and translated into English or vice versa. Automatic paragraph alignments between Chinese and English are available for these novels. The number of words in each Chinese passage in parallel passages is limited in the range of $[120, 600]$, to make the passage not too short or too long for crowdsourced workers to construct questions. As some of bilingual parallel passages are very difficult to understand and create parallel questions, we need to consider the appropriateness of these passages. The following rules are used to select bilingual parallel passages:
\begin{itemize}
    \item The Chinese passage shall not contain poetry, couplets, or classical Chinese words/phrases.
    \item The passage shall not contain too much dialogue, where speakers are difficult to recognize without global context.
    \item The Chinese passage shall not contain full-passage description of Kongfu fighting. Such descriptions on fighting are hard to translate in a direct way into a target language (i.e., English in this dataset).
    \item In order to ensure the correct alignments of Chinese and English passages, if an English passage has over 10 words fewer than its Chinese counterpart, such automatically aligned bilingual passages shall not be selected as they are normally not translations of each other.
\end{itemize}

Following these selection rules, we finally collect 3,667 bilingual parallel passages. Table \ref{table2} provides the number of selected passages from each novel.

\subsection{Question-Answer Pair Crowdsourcing}
We then ask our bilingual crowdsourced workers to create questions and answers on these collected passages. We develop a crowdsourcing annotation system and 150 bilingual workers, 3 bilingual reviewers and 1 expert participate in the data annotation process. The collected bilingual parallel passages are divided into 150 groups and randomly assigned to the bilingual workers. They will create bilingual parallel questions and find corresponding answers to the questions after reading the bilingual parallel passages. Particularly, we encourage workers to create questions according to the following rules:
\begin{itemize}
    \item For each parallel passage, at least three bilingual question-answer pairs are to be created.
    \item If the answers in Chinese and English are not parallel (i.e., not translations of each other), the corresponding questions shall be deleted and new questions shall be created. 
    \item Answers have to be consecutive spans in passages.
    \item If possible, questions of how and why are preferred.
    \item It is not suggested to directly copy words from passages for creating questions.
\end{itemize}

In order to guarantee the quality of created question-answer pairs, we use a strict quality control procedure during data annotation. In particular, 30\% annotated data from each group are randomly sampled and passed to the three reviewers who will review all answers created by the workers and correct answers if they consider they are wrong. Then, 5\% of the reviewed data will be further sampled from each reviewer. The sampled data will be checked again by the expert. If the accuracy is lower than 95\%, the corresponding workers and reviewers need to revise the answers again. This quality control loop is executed three times. 


At last, we collect 14,668 question-answer pairs along with their corresponding passages. We randomly partition the annotated data into a training set (with 11,668 QA pairs), a development set (1,500 QA pairs), and a test set (1,500 QA pairs).


\begin{table}[]
\small
\centering
\begin{tabular}{ccc}
\hline
Novels & \begin{tabular}[c]{@{}c@{}}\#Parallel Passages\end{tabular} & \begin{tabular}[c]{@{}c@{}}\#Avg EN/ZH \\ tokens\end{tabular} \\ \hline
\begin{tabular}[c]{@{}c@{}}\uppercase\expandafter{\romannumeral1}\\ \uppercase\expandafter{\romannumeral2}\\ \uppercase\expandafter{\romannumeral3}\\ \uppercase\expandafter{\romannumeral4}\\ \uppercase\expandafter{\romannumeral5}\\ \uppercase\expandafter{\romannumeral6}\end{tabular} & \begin{tabular}[c]{@{}c@{}}1,948\\ 75\\ 490\\ 245\\ 87\\ 822\end{tabular} & \begin{tabular}[c]{@{}c@{}}251.3/195.9\\ 223.3/194.6\\ 206.4/190.1\\ 175.1/188.9\\ 190.7/189.8\\ 202.4/212.3\end{tabular} \\
\midrule
Total & 3,667 & 227.3/198.2 \\ \hline
\end{tabular}
\caption{Statistics on the selected passages ( \uppercase\expandafter{\romannumeral1}: \textit{The Duke of the Mount Deer} /《鹿鼎记》, \uppercase\expandafter{\romannumeral2}: \textit{Demi-Gods and Semi-Devils} /《天龙八部》, \uppercase\expandafter{\romannumeral3}: \textit{The Three-Body Problem} /《三体》, \uppercase\expandafter{\romannumeral4}: \textit{The Great Gatsby} /《了不起的盖茨比》, \uppercase\expandafter{\romannumeral5}: \textit{The Old Man and the Sea} /《老人与海》, \uppercase\expandafter{\romannumeral6}: \textit{Harry Potter} /《哈利波特》).}
\label{table2}
\end{table}

\subsection{Multiple Answers Construction}

\label{ssec: multiple_Answers}

In order to make evaluation more robust, we ask crowdsourced workers to create at least two additional answers for each question in the development and test sets, similar to SQuAD \cite{rajpurkar2016squad}. But differently, we make the answers from the crowdsourced workers visible to each other, and encourage them to annotate different but reasonable answers. The reason for creating multiple amswers is that we often encounter situations where multiple answers are correct. Consider the following example from SQuAD\footnote{https://rajpurkar.github.io/SQuAD-explorer/explore/1.1/dev/}:

\textit{P: Official corporal punishment, often by caning, remains commonplace in schools in some Asian, African and Caribbean countries. For details of individual countries see School corporal punishment.}

\textit{Q:What countries is corporal punishment still a normal practice?}

The ground truth answers of the question are \textit{some Asian, African and Caribbean countries} \textbf{/} \textit{Asian, African and Caribbean}. The prediction of BERT \cite{devlin2019bert} ensemble model is \textit{Asian, African and Caribbean countries}. In fact, the machine-predicted result is also correct. However, it will be considered as a wrong answer if the exact match metric is used since the answer is not in the ground truth answer list. Such cases can be avoided if multiple reasonable answers are annotated.

\begin{figure}[tt]
\centering
\includegraphics[height=3.0in, width=3.0in]{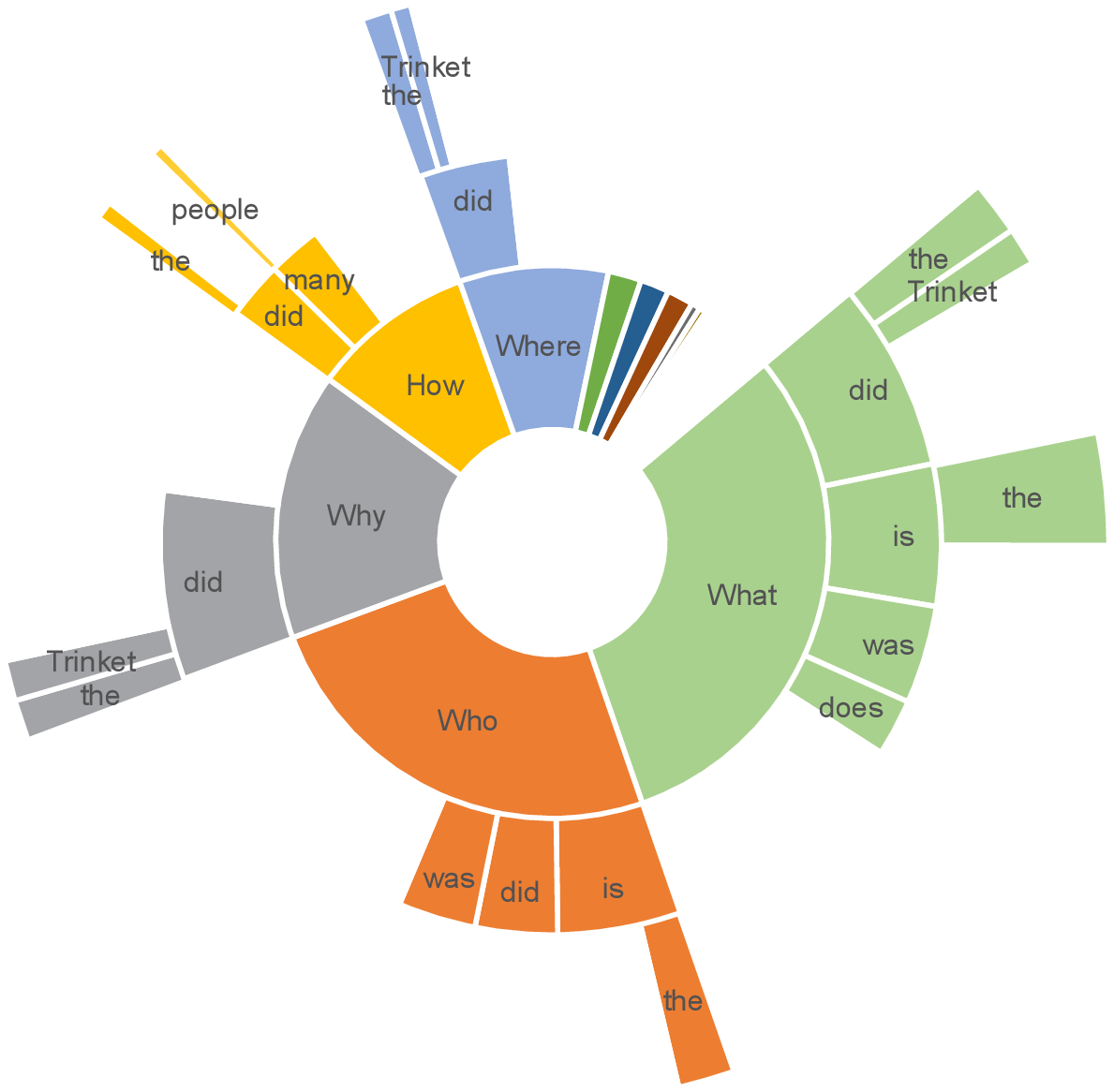}
\caption{Visualization of the distribution of trigram prefixes of questions in BiPaR.}
\label{question type}
\end{figure}

\section{Dataset Analysis}
 
In this section, we analyze the the types of questions and answers, the relationships of questions with passages as well as reading skills covered in BiPaR. Due to the bilingual parallelism of BiPaR, we choose the English part of the dataset for analyses. 

\subsection{Prefixes of Questions}

Figure \ref{question type} shows the distribution of trigram prefixes of questions in BiPaR. Unlike SQuAD \cite{rajpurkar2018know} where nearly half of questions are dominated by \textit{what} questions, the question type distribution in BiPaR is of better dispersion over multiple question types. In particular, BiPaR has 15.6\% \textit{why} and 9.5\% \textit{how} questions. Since in the novel texts, causality is usually not represented by explicit expressions such as ``why'', ``because'', and ``the reason for'', answering these questions in BiPaR requires the MRC models to understand implicit causality (Section \ref{ssec: li_phenomena}). The \textit{why} and \textit{how} questions, which account for considerable proportions, undoubtedly make BiPaR a very challenging MRC dataset.

\subsection{Answers Types}

\label{ssec: an_types}

We sample 100 examples from the development set, and present the types of answers in Table \ref{answer types}. As is shown, BiPaR covers a broad range of answer types, which matches our analysis on questions contribution. Moreover, we find that a large number of questions require some descriptive sentences to answer (37\%), which are generally complete sentences or summary statements, etc (See the third example in Table \ref{linguistic phenomena}). These answers are usually corresponding to \textit{what/why/how} questions, for instances:

\textit{1) \textbf{What} is Ding Yi doing?}

\textit{2) \textbf{Why} is fighting so much less fun?}

\textit{3) \textbf{How} did the protagonist treat her?}

\begin{table}[]
\small
\centering
\begin{tabular}{@{}lll@{}}
\toprule
Answer Type & \% & Examples \\ \midrule
\begin{tabular}[c]{@{}l@{}}Person\\ Location\\ Date\\ Verb phrase\\ Yes/No\\ Adjective\\ Event\\ Other proper noun\\ Common noun\\ Description\end{tabular} & \multicolumn{1}{c}{\begin{tabular}[c]{@{}c@{}}23\\ 13\\ 6\\ 2\\ 2\\ 3\\ 2\\ 4\\ 8\\ 37\end{tabular}} & \begin{tabular}[c]{@{}l@{}}Trinket\\ the floor\\ Saturday morning\\ wash her face\\ -\\ jade-green\\ Quidditch practice\\ a pinch of Floo powder\\ secret vault\\ wand backfired\end{tabular} \\ \bottomrule
\end{tabular}
\caption{Statistics on answer types in BiPaR.}
\label{answer types}
\end{table}

\begin{table*}[]
\centering
\resizebox{\textwidth}{!}{
\begin{tabular}{@{}clc@{}}
\toprule
Phenomenon & Example & \% \\ \midrule
\multicolumn{3}{c}{Relationship between a question and its passage} \\
\midrule
\begin{tabular}[c]{@{}c@{}}Lexical \\ match\end{tabular} & {\color[HTML]{000000} \begin{tabular}[c]{@{}l@{}}\textbf{P$_{en}$:} ... {\color{blue}{\textbf{Doublet}}} flew into the attack, flailing around her like the wind. She was too small to reach the bodies \\ of her enemies, ...jabbing the Vital Points on the riders' legs.\\ \textbf{P$_{zh}$:} ... {\color{blue}{\textbf{双儿}}}出手如风，只是敌人骑在马上，她身子又矮，打不到敌人，... 便戳中敌人腿上的穴道。\\ \textbf{Q$_{en}$}\textbf{-}\textbf{Q$_{zh}$:} Who did jab the Vital Points on the riders' legs? / 谁戳中敌人腿上的穴道？\end{tabular}} & 49 \\
Paraphrasing & \begin{tabular}[c]{@{}l@{}}\textbf{P$_{en}$:} ... Getting very difficult ter find anyone fer {\color{blue}{\textbf{the Dark Arts job}}}.\\ \textbf{P$_{zh}$:} ... 现在找一个{\color{blue}{\textbf{黑魔法防御术课老师}}}很困难，人们都不大想干，觉得这工作不吉利。\\ \textbf{Q$_{en}$}\textbf{-}\textbf{Q$_{zh}$:} What occupation do people avoid? / 人们不愿从事什么职业？\end{tabular} & 27 \\
Summary & \begin{tabular}[c]{@{}l@{}}\textbf{P$_{en}$:} ... As soon as he opened the door to Ding Yi's brand-new three-bedroom apartment, ... {\color{blue}{\textbf{The apartment}}} \\ {\color{blue}{\textbf{was unfinished, with only a few pieces of furniture and little decoration, and the huge living room seemed}}} \\ {\color{blue}{\textbf{very empty. The most eye-catching object was the pool table in the corner}}}.\\ \textbf{P$_{zh}$:} ... 推开丁仪那套崭新的三居室的房门，...
看到{\color{blue}{\textbf{房间还没怎么装修，也没什么家具和陈设，宽大的}}}\\ {\color{blue}{\textbf{客厅显得很空，最显眼的是客厅一角摆放的一张台球桌}}}。\\ \textbf{Q$_{en}$}\textbf{-}\textbf{Q$_{zh}$:} What does Ding Yi's three-bedroom look like now? / 丁仪的三居室现在是什么样？\end{tabular} & 24 \\
\midrule
\multicolumn{3}{c}{Reading comprehension skills required to answer questions} \\
\midrule
\begin{tabular}[c]{@{}c@{}}Coreference \\ resolution\end{tabular} & \begin{tabular}[c]{@{}l@{}}\textbf{P$_{en}$:} Trying hard to bear all this in mind, {\color{cyan}{\textbf{Harry}}} took a ... {\color{cyan}{\textbf{he}}} opened his mouth and immediately swallowed \\ a lot of hot ash.``D-{\color{blue}{\textbf{Dia-gon Alley}}}," {\color{cyan}{\textbf{he}}} coughed.\\ \textbf{P$_{zh}$:} {\color{magenta}{\textbf{哈利}}}拼命把这些都记在心里，... {\color{magenta}{\textbf{他}}}一张嘴，马上吸了一大口滚烫的烟灰。“对一{\color{blue}{\textbf{对角巷}}}。”\\ {\color{magenta}{\textbf{他}}}咳着说。\\ \textbf{Q$_{en}$}\textbf{-}\textbf{Q$_{zh}$:} Where did {\color{cyan}{\textbf{Harry}}} go? / {\color{magenta}{\textbf{哈利}}}去了哪儿？\end{tabular} & 31 \\
\begin{tabular}[c]{@{}c@{}}Multi-sentence \\ reasoning\end{tabular} & \begin{tabular}[c]{@{}l@{}}\textbf{P$_{en}$:} and heard a woman's voice cry out from within it: `Stop! Lay down your arms! We should all be friends \\ here!' ... The cart stopped in front of them, and out jumped—{\color{blue}{\textbf{Fang Yi}}}.\\ \textbf{P$_{zh}$:} 车中一个女子声音叫道：“是自己人，别动手！” ... 小车驶别跟前，车中跃出一人，正是{\color{blue}{\textbf{方怡}}}。\\ \textbf{Q$_{en}$}\textbf{-}\textbf{Q$_{zh}$:} Who did stop the conflict? / 是谁制止了冲突？\end{tabular} & 32 \\
\begin{tabular}[c]{@{}c@{}}Implicit \\ causality\end{tabular} & \begin{tabular}[c]{@{}l@{}}\textbf{P$_{en}$:} Harry, however, was shaken awake several hours earlier than he would have liked by Oliver Wood, Captain \\ of the Gryffindor Quidditch team.``Whassamatter?" said Harry groggily. ``{\color{blue}{\textbf{Quidditch practice}}}" said Wood.\\ \textbf{P$_{zh}$:} 哈利一早就被格兰芬多魁地奇队队长奥利弗伍德摇醒了，他本来还想再睡几个小时的。“什一什么事？”\\ 哈利迷迷糊糊地说。“{\color{blue}{\textbf{魁地奇训练}}}”伍德说。\\ \textbf{Q$_{en}$}\textbf{-}\textbf{Q$_{zh}$:} Why did Oliver Wood shake Harry awake? / 奥利弗伍德为什么要摇醒哈利？\end{tabular} & 17 \\ \bottomrule
\end{tabular}}
\caption{Question categories and reading comprehension skills covered in BiPaR. The blue indicates the answer, and other colors indicate coreference resolution.}
\label{linguistic phenomena}
\end{table*}

\subsection{Relationships of Questions with Passages and Reading Comprehension Skills }

\label{ssec: li_phenomena}

In order to assess how difficult to answer questions in BiPaR, we further analyze the relationships between questions and corresponding passages as well as reading comprehension skills required to detect answers for BiPaR questions. We sample 100 samples from the development set and annotate them with various reasoning phenomena as shown in Table \ref{linguistic phenomena}.

Inspired by \citet{reddy2019coqa}, we group questions into several categories in terms of their relationships with passages. If a question contains more than one content word that appears in the passage, we label it as \textit{lexical match}. These account for 49.0\% of all the questions. One might think that if a lot of words in a question overlap with those in a passage, the answer may be easily detected from the matched sentence in the passage. However, this is not the case at all in BiPaR. As the first example in Table \ref{linguistic phenomena} shows, the question is almost the same to the sentence in passage. However, the answer is far away from the matched sentence. Actually, correctly answering this question requires very complicated reading comprehension skills, such as multi-sentence reasoning, ellipsis/co-reference resolution, etc. We've found 43 examples of this case among the 49 lexical-match samples. 

If there is no lexical match between a question and the corresponding passage but we can find a semantically matched sentence to the question from the passage, we regard this case as \textit{paraphrasing}. Such questions account for 27.0\% of the questions. Interestingly, we find questions in BiPaR that are not found in other datasets. We refer to these questions as \textit{summary} questions, which account for 24\% of the sampled questions. In order to answer these questions, MRC models need to read the entire passage to detect summative statements. Examples of the summary questions are:

\textit{1) What is the situation of the Old Majesty?}

\textit{2) What features does Oboi's bedroom show?}

In addition, we also analyze the reading comprehension skills required to answer questions. We find that \textit{coreference resolution}, \textit{multi-sentence reasoning} and \textit{implicit causality understanding} frequently appear in answering questions in BiPaR. What deserves our special attention here is the implicit causality, which rarely appears in other datasets. For some questions, it is crucial to understand causality that is not represented by explicit expressions such as ``why'', ``because'', and ``the reason for''. As demonstrated in the last example in Table \ref{linguistic phenomena}, to correctly answer the question, we must understand the implicit causality: \textit{Quidditch practice} $\longrightarrow$ \textit{Harry, however, was shaken awake several hours earlier than he would have liked by Oliver Wood, Captain of the Gryffindor Quidditch team}.

\section{MRC Task Formulation on BiPaR}

\label{ssec: tasks}

With aligned passage-question-answer triples, we can define three MRC tasks (monolingual, multilingual and cross-lingual) with seven different forms on this dataset, as demonstrated in Figure 1.  Since our goal is to provide benchmark results on this new dataset, we either directly train state-of-the-art MRC models on these tasks or use a straightforward way to adapt existing approaches to the defined tasks in this paper. We leave new approaches, especially those for multilingual and cross-lingual MRC to our future work.\\

\noindent\textbf{Monolingual MRC}: ($P_{en}$, $Q_{en}$, $A_{en}$) or ($P_{zh}$, $Q_{zh}$, $A_{zh}$). With these two monolingual MRC forms, we can investigate the performance variation of the same MRC model trained on two different languages with equivalent training instances. In our experiments, we directly train off-the-shelf MRC models on the two monolingual tasks to evaluate their performance on Chinese and English.\\

\noindent\textbf{Multilingual MRC}: ($P_{en}$, $Q_{en}$, $A_{en}$, $P_{zh}$, $Q_{zh}$, $A_{zh}$). Similar to multilingual neural machine translation \cite{johnson2017google}, we can build a single MRC model to handle MRC of multiple languages on BiPaR. In our benchmark test, we directly mix training instances of the two languages into a single training set. Correspondingly, the two vocabularies are also combined into one vocabulary for both languages. After that, we train MRC models on this language-mixed dataset to endow them with the multilingual comprehension capacity.\\

\noindent\textbf{Cross-lingual MRC}: The first two forms of cross-lingual MRC are ($P_{en}$, $Q_{zh}$, $A_{en}$) or ($P_{zh}$, $Q_{en}$, $A_{zh}$), in which we use questions in one language to extract answers from passages written in another language. The other two forms are ($P_{en}$, $P_{zh}$, $Q_{zh}$, $A_{zh}$, $A_{en}$) or ($P_{zh}$, $P_{en}$, $Q_{en}$, $A_{en}$, $A_{zh}$), in which we use questions written in one language to extract answers from passages written in multiple languages. For the first two forms of cross-lingual MRC, we use Google Translate\footnote{https://translate.google.com/} to translate questions into the language of passages, and then treat them as a monolingual MRC task. For the second two forms of cross-lingual MRC, such as ($P_{zh}$, $P_{en}$, $Q_{en}$, $A_{en}$, $A_{zh}$), we first obtain $A_{en}$ through a monolingual MRC model, then use the word alignment tool fast\_align\footnote{https://github.com/clab/fast\_align} to obtain the aligned $A_{zh}$ from $P_{zh}$.  Alternative approaches that do not rely on machine translation or word alignments for the cross-lingual MRC tasks are to directly build cross-lingual MRC models on language-mixed training instances constructed from BiPaR or to explore multi-task learning on multiple languages \cite{dong2015multi}.

\begin{table*}[]
\centering
\resizebox{\textwidth}{!}{
\begin{tabular}{@{}lccccccc@{}}
\toprule
 & \multicolumn{2}{c}{Monolingual MRC} & Multilingual MRC & \multicolumn{4}{c}{Cross-lingual MRC} \\ \midrule
 & 1 & 2 & 3 & 4 & 5 & 6 & 7 \\
 \midrule
\multicolumn{8}{c}{Development set} \\
\midrule
DrQA & 29.87/43.47 & 36.60/52.90 & 31.47/44.93 36.68/54.03 & 27.80/38.94 & 28.47/43.65 & 8.07/19.80 & 6.27/19.79 \\
BERT\_base & 41.67/56.23 & \textbf{52.53/67.65} & 42.33/55.49 49.00/63.99 & 36.27/49.98 & \textbf{41.93/55.66} & 8.00/24.37 & 8.60/22.51 \\
BERT\_large & \textbf{44.47/58.94} & - & - & \textbf{40.40/53.28} & - & 7.87/24.60 & - \\
\midrule
\multicolumn{8}{c}{Test set} \\
\midrule
DrQA & 27.00/39.29 & 37.40/53.11 & 28.00/42.49 36.60/53.34 & 21.93/34.45 & 27.53/41.08 & 7.00/18.63 & 4.07/16.64 \\ 
BERT\_base & 41.40/55.03 & \textbf{48.87/64.09} & 38.33/51.20 49.00/64.06 & 32.80/46.36 & \textbf{39.87/53.10} & 5.73/21.08 & 7.67/20.69 \\
BERT\_large & \textbf{42.53/56.48} & - & - & \textbf{37.53/51.51} & - & 5.60/22.29 & - \\
\midrule
Human & 80.50/91.93 & 81.50/92.12 &  &  &  &  &  \\ \bottomrule
\end{tabular}}
\caption{Results (EM/F1 score) of models and humans on the development and the test data  of BiPaR. 1-7 indicate the seven different MRC tasks on BiPaR:  ($P_{en}$, $Q_{en}$, $A_{en}$), ($P_{zh}$, $Q_{zh}$, $A_{zh}$), ($P_{en}$, $Q_{en}$, $A_{en}$, $P_{zh}$, $Q_{zh}$, $A_{zh}$), ($P_{en}$, $Q_{zh}$, $A_{en}$), ($P_{zh}$, $Q_{en}$, $A_{zh}$), ($P_{en}$, $P_{zh}$, $Q_{zh}$, $A_{zh}$, $A_{en}$), ($P_{zh}$, $P_{en}$, $Q_{en}$, $A_{en}$, $A_{zh}$). For the 6th and 7th task, we mainly explored the word-alignment method. Hence, EM and F1 scores were evaluated on $A_{en}$ or $A_{zh}$.}
\label{results}
\end{table*}

\section{Experiments}

 We carried out experiments with state-of-the-art MRC models on BiPaR to provide machine results for these 7 MRC tasks defined above. We also provide human performance on the monolingual tasks and demonstrate the performance trajectory of human and machine in answering BiPaR questions. 

\subsection{Evaluation Metric}

Like evaluations on other extraction-based datasets, we used EM and F1 to evaluate model accuracy on BiPaR. Particularly, we used the evaluation program of SQuAD1.1 for the English dataset in BiPaR, and the evaluation program\footnote{https://github.com/ymcui/cmrc2018} of CMRC2018 for the Chinese dataset in BiPaR.

\subsection{Human Performance Evaluation}

In order to assess human performance on BiPaR, we hired three other bilingual crowdsourced workers to independently answer questions (both Chinese and English) on the test set which contains three answers per question as described in Section \ref{ssec: multiple_Answers}. We then calculated the average results of the three human workers as the final human performance on this dataset, which are shown in Table \ref{results}.

\subsection{Baseline models}

We adapted the following state-of-the-art models to the dataset and MRC tasks as described in Section 4.

\noindent\textbf{DrQA\footnote{https://github.com/hitvoice/DrQA}:} DrQA \cite{chen-etal-2017-reading} is a simple but effective neural network model for reading comprehension.

\noindent\textbf{BERT\footnote{https://github.com/huggingface/pytorch-pretrained-BERT}:} BERT \cite{devlin2019bert} is a strong method for pre-training language representations, which obtains the state-of-the-art results on many reading comprehension datasets. We used the multilingual model of BERT trained on multiple languages for the evaluation of our multilingual MRC task. 

\subsection{Experimental Setup}

All the baselines were tested using their default hyper-parameters except BERT. We only changed the batch size to 8 for BERT$\_$base and 6 for BERT$\_$large due to the memory limit of our GPUs\footnote{The original batch sizes used in BERT$\_$base/BERT$\_$large are 12/24.}. We used spaCy\footnote{https://spacy.io/} to tokenize sentences and generate part-of-speech and named entity tags that were used to train the DrQA model. We downloaded Chinese models\footnote{https://github.com/howl-anderson/Chinese\_models\_for\_SpaCy} for spaCy to preprocess Chinese datasets. The 300-dimensional Glove word embeddings trained from 840B Web crawl data \cite{pennington2014glove} were used as our pre-trained English word embeddings while the 300-dimensional SGNS word embeddings trained from mixed-sources data \cite{li2018analogical} as our pre-trained Chinese word embeddings.





\subsection{Evaluation Results}

Table \ref{results} presents the results of the models on the development and the test data. As BERT$\_$large is currently not available for Chinese, results on the tasks that need to use BERT$\_$large on Chinese are not provided.\\

\noindent\textbf{BERT vs. Human:} In the monolingual MRC task, the English BERT$\_$large model achieves an F1 score of 56.5\%, which is 35.4 points behind human performance (91.9\%), and the Chinese BERT$\_$base model achieves an F1 score of 64.1\%, which is 28.0 points behind human performance (92.1\%), indicating that these tasks are difficult to accomplish with current state-of-the-art models. We also tested the English BERT$\_$large model on a subset of the SQuAD data, which contains the same number of training instances as BiPaR. The F1 score is 86.5\%, much more higher than that on BiPaR. This further suggests that BiPaR provides a very challenging dataset for MRC.\\

\noindent\textbf{English vs. Chinese:} On the monolingual and cross-lingual tasks, Chinese results are almost better than the English results of the same MRC model. This does not mean that Chinese MRC is easier than English. One possible reason for this may be that novels originally written in Chinese contribute to 68.5\% passages of BiPaR. In the future, we plan to make the dataset more balanced between the two languages. \\

\noindent\textbf{Monolingual vs. Multilingual:} For the DrQA model, we observe that the multilingual training significantly improves the performance on English comparing with the monolingual training with only the English dataset. However, we do not observe this trend on BERT. This suggests that more should be explored on the multilingual MRC setting. We believe that BiPaR opens a door to new MRC approaches that are devoted to using a single model to handle MRC on multiple languages.\\

\noindent\textbf{Monolingual vs. Cross-lingual:} The two simple strategies via machine translation and word alignments for the cross-lingual MRC perform very bad on the four forms of cross-lingual tasks compared to the monolingual task. For ($P_{en}$, $Q_{zh}$, $A_{en}$) and ($P_{zh}$, $Q_{en}$, $A_{zh}$), the major problem is the low-quality translations of questions, especially for questions from martial arts novels. For the other two cross-lingual tasks, word alignment errors directly result in wrong answers found in passages of the other language.

\subsection{Analysis of BERT and Human in Answering Different BiPaR Questions}

Table \ref{error analysis} presents a fine-grained comparison analysis of BERT$\_$base and human results on the English and Chinese monolingual task in terms of both answer types and question categories defined in Section \ref{ssec: li_phenomena}. We observe that humans have absolute advantages over machines in all answer types and reasoning phenomena. However, humans exhibit different capabilities on answering different questions.  They perform worse on answering paraphrasing questions, questions with description answers and questions requiring reading comprehension skills of multi-sentence reasoning than answering other questions. 

The performance of BERT$\_$base on questions with description answers is much worse than other questions, e.g., questions with person/location answers. As described in Section \ref{ssec: an_types}, descriptive answers are often complete sentences or summary statements, which are very long and difficult for machines  to detect. 

In terms of the relationships between questions and passages, it is clearly observed that the BERT MRC model achieves better performance on answering lexical-match questions than answering paraphrasing and summary questions. In BiPaR, we have more than 50\% questions that are paraphrasing and summary questions as described in Section \ref{ssec: li_phenomena}. Answering them requires a deep understanding of both questions and passages.

As described in Section \ref{ssec: li_phenomena}, BiPaR produces questions that involve higher-order reading comprehension skills, such as co-reference resolution, multi-sentence reasoning, and understanding of implicit causality. It can be seen from Table \ref{error analysis} that the BERT model is worse on multi-sentence reasoning and implicit causality than co-reference resolution.

\begin{table}[]
\small
\centering
\begin{tabular}{@{}lcc@{}}
\toprule
Type & BERT\_base & Human \\ \midrule
\multicolumn{3}{c}{Answer Type} \\
\midrule
\begin{tabular}[c]{@{}l@{}}Non-description\\ Description\end{tabular} & \begin{tabular}[c]{@{}c@{}}61.97/63.22\\ 44.77/54.34\end{tabular} & \begin{tabular}[c]{@{}c@{}}91.72/95.10\\ 91.87/94.72\end{tabular} \\
\midrule
\multicolumn{3}{c}{Question Category} \\
\midrule
\begin{tabular}[c]{@{}l@{}}Lexical Match\\ Paraphrasing\\ Summary\end{tabular} & \begin{tabular}[c]{@{}c@{}}58.69/69.01\\ 54.34/49.84\\ 50.73/52.76\end{tabular} & \begin{tabular}[c]{@{}c@{}}96.09/96.73\\ 85.49/91.36\\ 90.03/95.39\end{tabular} \\
\midrule
\begin{tabular}[c]{@{}l@{}}Coreference res.\\ Multi-sentence rea.\\ Implicit causality\end{tabular} & \begin{tabular}[c]{@{}c@{}}52.35/58.97\\ 44.91/45.81\\ 39.31/57.63\end{tabular} & \begin{tabular}[c]{@{}c@{}}91.78/98.11\\ 87.22/95.31\\ 92.54/91.28\end{tabular} \\ \bottomrule
\end{tabular}
\caption{Fine-grained results in terms of different answer types and question categories on the monolingual task. The left side of the slash is the F1 score on the English data, while the right is on Chinese. All F1 scores are calculated on the 100 questions as described in Section \ref{ssec: li_phenomena}.}
\label{error analysis}
\end{table}

\section{Conclusion and Future Work}

In this paper, we have presented the BiPaR, a bilingual parallel machine reading comprehension dataset on novels. From bilingual parallel passages of Chinese and English novels, we manually created diversified parallel questions and answers of different types via crowdsourced workers with a multi-layer quality control  system. Although BiPaR is an extractive MRC dataset, in-depth analyses demonstrate that the dataset is very challenging for state-of-the-art MRC models (performing far behind human) as reading comprehension skills of co-reference resolution, inter-sentential reasoning are needed to answer BiPaR questions. We further define seven types of MRC tasks supported by BiPaR and build baseline models of monolingual, multilingual and cross-lingual MRC on BiPaR. 

BiPaR can be extended in several ways. First, we would like to create more parallel triples by adding more novels to make instances more balanced between the two languages. Second, we want to create questions with non-extractive answers. Third, we are also interested in adding multi-passage questions or questions based on the entire novels to BiPaR.

\section*{Acknowledgments}

The present research was supported by the National Natural Science Foundation of China (Grant No. 61622209). We would like to thank the anonymous reviewers for their insightful comments. \\



\bibliography{emnlp-ijcnlp-2019}
\bibliographystyle{acl_natbib}

\clearpage

\appendix

\end{CJK}
\end{document}